\documentclass[runningheads]{llncs}

\usepackage{eccv}

\usepackage[dvipsnames]{xcolor}

\usepackage{booktabs}
\usepackage{multicol}
\usepackage{multirow}
\usepackage{colortbl}
\usepackage{xcolor}
\usepackage{caption}
\usepackage{tabularray}
\usepackage{rotating}
\usepackage{makecell}
\usepackage{tabularx}
\usepackage{float}
\usepackage{adjustbox}
\usepackage{anyfontsize}
\usepackage{mathtools}
\usepackage{amsmath}
\usepackage{indentfirst}
\usepackage{float}

\usepackage{amssymb}
\usepackage{pifont}
\newcommand{\cmark}{\ding{51}}%
\usepackage{microtype}

    \makeatletter
\def\@fnsymbol#1{\ensuremath{\ifcase#1\or \dagger\or *\or
   \mathsection\or \mathparagraph\or \|\or **\or \dagger\dagger
   \or \ddagger\ddagger \else\@ctrerr\fi}}
    \makeatother

\usepackage{eccvabbrv}

\usepackage{graphicx}
\usepackage{booktabs}

\usepackage[accsupp]{axessibility}  

\usepackage{hyperref}

\usepackage{orcidlink}

\begin{document}

\title{Generative Dataset Distillation using Min-Max Diffusion Model} 

\titlerunning{Dataset Distillation with Diffusion}

\author{Junqiao Fan\inst{1}\orcidlink{0000-0002-8465-5447}\index{Fan, Junqiao} \and
Yunjiao Zhou\inst{1}\orcidlink{0009-0009-7515-1739}\index{Zhou, Yunjiao} \and  Min Chang Jordan Ren\inst{2}\index{Min Chang Jordan Ren}\thanks{M. Ren is a research intern at Nanyang Technological University.}\and
Jianfei Yang\inst{1,2}\thanks{J. Yang is the corresponding author (jianfei.yang@ntu.edu.sg).}\orcidlink{0000-0002-8075-0439}\index{Yang, Jianfei}}

\authorrunning{J.~Fan et al.}

\institute{School of Electrical and Electronic Engineering\\
\and
School of Mechanical and Aerospace Engineering\\
Nanyang Technological University, Singapore\\
\email{\{fanj0019, yunjiao001\}@e.ntu.edu.sg,\\ jordan9ren8@gmail.com, jianfei.yang@ntu.edu.sg}}

\maketitle

\begin{abstract}
  In this paper, we address the problem of generative dataset distillation that utilizes generative models to synthesize images. The generator may produce any number of images under a preserved evaluation time. In this work, we leverage the popular diffusion model as the generator to compute a surrogate dataset, boosted by a min-max loss to control the dataset's diversity and representativeness during training. However, the diffusion model is time-consuming when generating images, as it requires an iterative generation process. We observe a critical trade-off between the number of image samples and the image quality controlled by the diffusion steps and propose Diffusion Step Reduction to achieve optimal performance. This paper details our comprehensive method and its performance. Our model achieved $2^{nd}$ place in the generative track of \href{https://www.dd-challenge.com/#/}{The First Dataset Distillation Challenge of ECCV2024}, demonstrating its superior performance. 
  \keywords{Dataset distillation \and Diffusion model \and Generative model}
\end{abstract}

\section{Introduction}
\label{sec:introduction}

The success of ChatGPT has demonstrated the power of ultra-large-scale and high-quality data. However, the high demand for data imposes challenges for storage and computation resources~\cite{he2016deep, dosovitskiy2020image}. Therefore, dataset distillation (DD) is proposed to condense rich information from a large-scale dataset into a surrogate one, achieving comparable training performance~\cite{wang2018dataset, cui2022dc, kim2022dataset, zhao2020dataset}.

In this paper, we focus on the problem of generative DD in \href{https://www.dd-challenge.com/#/}{The First Dataset Distillation Challenge of ECCV2024}. Generative models~\cite{oussidi2018deep, kingma2021variational, goodfellow2014generative,
song2021score} are leveraged to distillate on image datasets (e.g., ImageNet~\cite{deng2009imagenet}), which generates image samples for the surrogate dataset. Different from the traditional Images Per Class (IPC) setting that produces a fixed amount of training samples~\cite{martinez2003recognizing, howard2019smaller, shahinfar2020many, liu2023dream}, generative DD allows the generator to produce any number of images given a preserved evaluation time. Specifically in the challenge, 10 minutes are given for the generator to produce image samples on a single RTX 4090 GPU. This setting can ensure a fair comparison between different types of models, e.g. GAN~\cite{creswell2018generative, goodfellow2020generative, karras2019style} and diffusion models~\cite{dhariwal2021diffusion, ho2020denoising, nichol2021improved}. 

Diffusion models have recently surpassed GAN in high-quality image generation tasks. They can model the distribution of the given training dataset, demonstrating the potential for generative DD tasks by generating high-quality representative images~\cite{gu2024efficient, ho2020denoising, dhariwal2021diffusion, bengio2013representation}. This paper presents a generative DD method using a diffusion model. However, the diffusion-based method confronts two major challenges as it is more time-consuming than GAN-based methods due to its iterative generative process. Firstly, the representativeness and diversity of the surrogate dataset should be ensured given a fixed amount of time. The generated samples should effectively capture the major features of the original datasets. In the meantime, they should generate all possible image samples to ensure diversity. Secondly, the number of generated samples in the surrogate dataset should be increased within the given time. Intuitively, more training samples would lead to better training performance. 

To solve the abovementioned two challenges, we first leverage two min-max training losses to ensure the representativeness and diversity of the surrogate dataset respectively. Then, we propose Diffusion Steps Reduction (DSR) to carefully control the number of diffusion iterations during image generation, optimizing the trade-off between the quality and quantity of the image samples. Comprehensive experiments demonstrate that our method produces representative results, ranking second in the generative track of \href{https://www.dd-challenge.com/#/}{The First Dataset Distillation Challenge of ECCV2024}. Additionally, it achieves a strong trade-off between the quality and quantity of generated images, optimizing performance for the given evaluation constraints.

\section{Methodology}
\label{sec:methodology}
\begin{figure*}[!t]
\centering
\includegraphics[width=1.\linewidth]{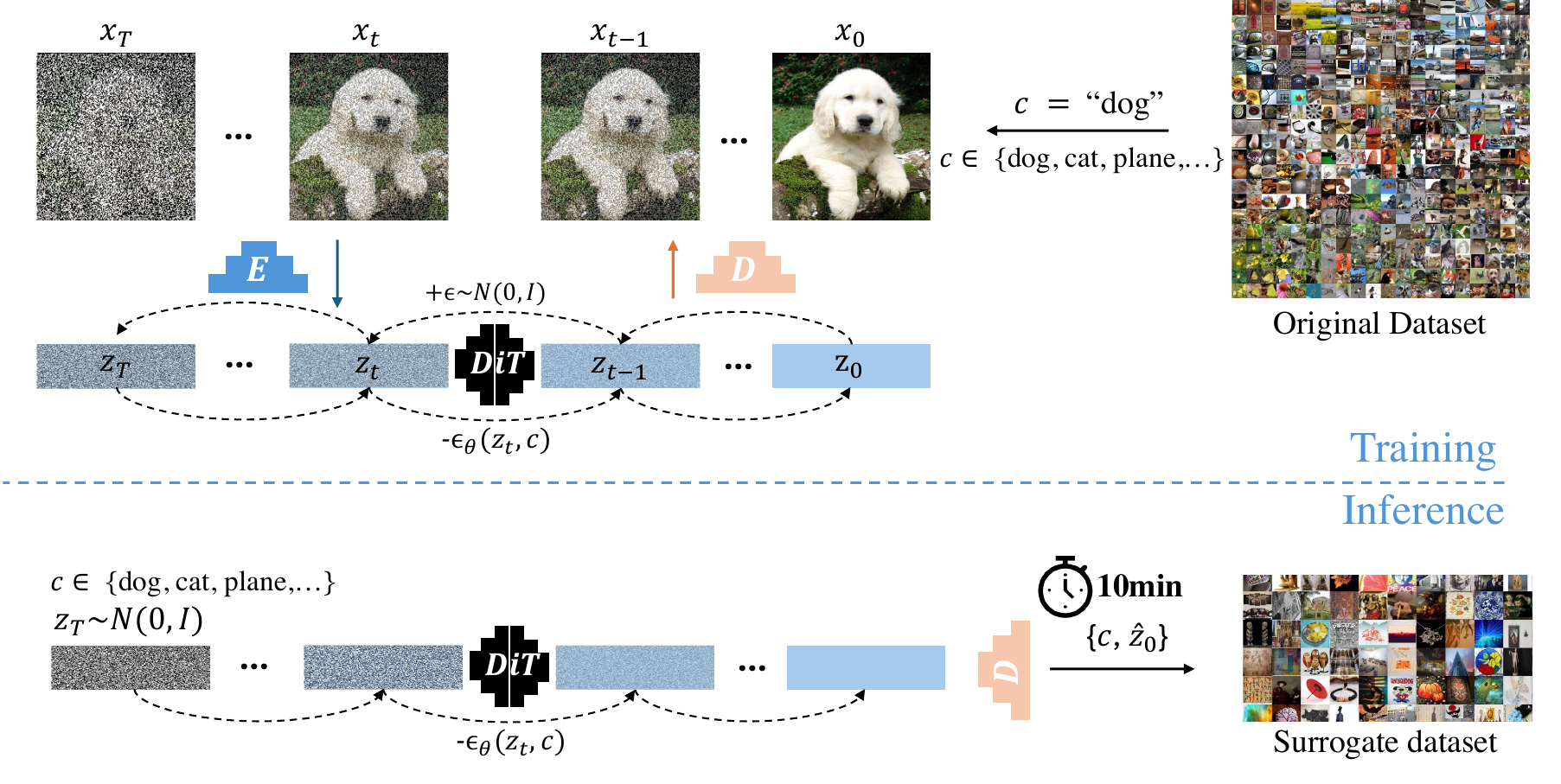}
\caption{During training, the diffusion model learns a dataset distribution by gradually adding Gaussian noise to images and reversing back. The model is trained by $\mathcal{L}_{train} = \mathcal{L}_{diff} + \lambda_r*\mathcal{L}_r + \lambda_d*\mathcal{L}_d$. During inference, the trained model utilizes 10 mins to generate samples for the surrogate dataset, where DSR is utilized to accelerate image generation.}
\label{system architecture}
\end{figure*}


\subsection{Diffusion-based Dataset Distillation}
Diffusion models can learn a dataset distribution by gradually adding Gaussian noise to images and reversing back. We leverage the latent diffusion model (LDM)~\cite{rombach2022high} for more efficient implementation, which learns the distribution of the feature space. Given an image $x$, an encoder $E$ first transforms the image into the latent feature representation $z = E(x)$ and a decoder $D$ learns to reconstruct the latent feature back to the image space $\hat{x} = D(z)$. In the forward process, Gaussian noise $\varepsilon \in N(0, I)$ is gradually added to the original latent code:~$z_t=\sqrt{\gamma_t} z_0+\sqrt{1-\gamma_t} \varepsilon$, where $\gamma_t$ indicates noise scale. In the reverse process, the diffusion model is trained to predict and reverse the noise $\hat{\varepsilon}_\theta(z_t, c)$, guided by a conditioning vector $c$ indicating the class label. The model is trained by the diffusion loss:

\begin{equation}
    \mathcal{L}_{\text {diff}}=|| \varepsilon - \hat{\varepsilon}_\theta(z_t, c)\|^2_2 \\,
\label{eq:diff2-supp}
\end{equation}
where $\theta$ is the diffusion network parameters.

\subsection{Min-max Training Loss}
Following \cite{gu2024efficient}, two min-max training losses, are designed for improving the representativeness and diversity of samples generated by the diffusion model. 

\subsubsection{Representativeness.} Firstly, for a specific class, the generated samples should be the most representative images, which capture the major features from all possible samples. Therefore, we first store the real image features from adjacent mini-batch into a real-feature set $M = \{z_m\}_{m=1}^{N_M}$. Then, given a predicted feature representation $\hat{z}_\theta(z_t, c) = z_t - \varepsilon_\theta(z_t, c)$, a min-max representativeness loss $\mathcal{L}_r$ is proposed to pull close the least similar sample pairs within the real-feature set:
\begin{equation}
    \mathcal{L}_r=\arg \max _\theta \min _{m \in\left[N_M\right]} \sigma\left(\hat{{z}}_\theta\left({z}_t, {c}\right), {z}_m\right),
\label{eq:l_r}
\end{equation}
where $\sigma(\cdot)$ denotes the cosine similarity.

\subsubsection{Diversity.} The generated samples should also produce diverse images for a surrogate dataset. Therefore, we first store the synthesized image samples from adjacent mini-batch into a synthesized-feature set $D = \{z_d\}_{d=1}^{N_D}$. Then, given the predicted feature representation $\hat{z}_\theta(z_t, c)$, another min-max diversity loss $\mathcal{L}_d$ is proposed to push away the most similar pairs within the synthesized set:
\begin{equation}
    \mathcal{L}_d=\arg \min _\theta \max _{d \in\left[N_D\right]} \sigma\left(\hat{{z}}_\theta\left(z_t, {c}\right), {z}_d\right).
\label{eq:l_d}
\end{equation} The overall learning objective is as follows:
\begin{equation}
    \mathcal{L}_{train} = \mathcal{L}_{diff} + \lambda_r*\mathcal{L}_r + \lambda_d*\mathcal{L}_d,
\label{eq:all}
\end{equation}
where $\lambda_r$ and $\lambda_d$ are the weighting parameters.

\subsection{Diffusion Step Reduction}
Given a trained generative model, increasing the number of image samples during inference could lead to better results. However, the diffusion model is more time-consuming than GAN when producing the surrogate datasets, due to the iterative diffusion steps. Therefore, we propose a strategy called Diffusion Step Reduction (RSD), which reduces the number of iterations during surrogate dataset generation. RSD aims to increase the number of generated image samples by saving the generation time, while sacrificing image quality. Nevertheless, by carefully selecting the number of diffusion steps, this strategy can maintain desirable image quality and effectively improve performance.

\section{Experiments}
\label{sec:experiments}

\begin{figure}[!t]
\centering
\includegraphics[width=1.\linewidth]{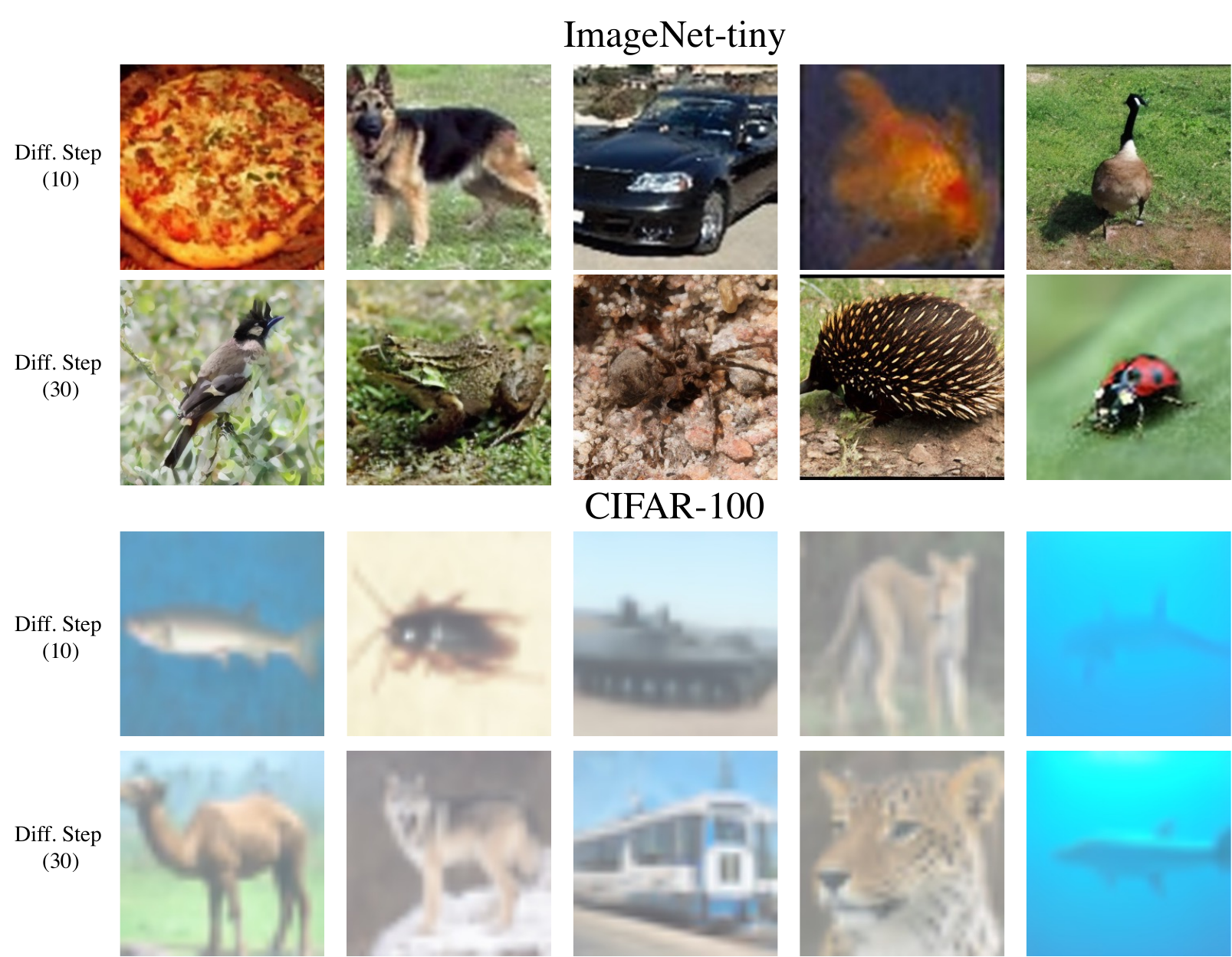}
\caption{Qualitative visualization of the surrogate dataset on ImageNet-tiny and CIFAR-100, where 10 and 30 diffusion iterations are performed to generate images.}
\label{exp1} 
\end{figure}

\subsection{Experiment Setup}
We follow the experiment settings of \href{https://www.dd-challenge.com/#/}{The First Dataset Distillation Challenge of ECCV2024} using an RTX 3090 GPU. Our implemented methods are tested on ImageNet-Tiny~\cite{deng2009imagenet} (200 classes) and CIFAR-100~\cite{Krizhevsky09learningmultiple} (100 classes) datasets respectively. Firstly, 10 minutes are given for the generative models to generate the surrogate datasets. Then, an evaluation classifier, ConvNetD3-W128, is trained using the surrogate datasets for 1000 epochs. The learning rate, momentum, and weight decay are set to 0.01, 0.9, and 0.0005 respectively. The accuracy of image classification is recorded for comparison. Moreover, we utilize Image Per Plass (IPC) to quantify the speed of image sample generation, which is calculated by $IPC = \frac{\# of samples}{\# of classes}$.

\subsection{Implementation Details}
\label{Implementation Details}

We adopt pretrained DiT~\cite{peebles2023scalable} as the baseline diffusion model and set the image size as $256\times256$. As different datasets have different image resolutions, the images are re-scaled to $256\times256$ for training, and the generated images will be scaled back to their original resolution for inference. We set $N_M=N_D=64$ for our the real feature set $M$ and synthesized feature set $N$. $\lambda_r=1e-3$ and $\lambda_d=2e-3$ are set for calculating the training loss in Eq.\ref{eq:all}. During training, all implemented methods are trained for 8 epochs with a batch size of 8. The Adam algorithm~\cite{kingma2014adam} is used for optimization, with the learning rate as $1e-4$. During inference, we set the number of diffusion steps as 10 for generating image samples in DSR.

\begin{table}[!t]
\centering
\caption{Quantitative generative dataset distillation performance on ImageNet-tiny and CIFAR-100 datasets. Bold indicates the best}

\renewcommand{\arraystretch}{1.5}
\resizebox{.65\textwidth}{!}{

    \begin{tabular}{p{13em}ccp{1em}cc}
    \toprule
    Methods & \multicolumn{2}{c}{ImageNet-tiny} &       & \multicolumn{2}{c}{CIFAR-100} \\
\cmidrule{2-3}\cmidrule{5-6}          & IPC & Accuracy &       & IPC & Accuracy \\
    \midrule
    DiT   &    $3$   &    $3.50 \scriptstyle{\pm 0.16}$   &   &  $6$  &    $5.63 \scriptstyle{\pm 0.47}$      \\
    DiT + Min-Max &    $3$   &   $3.53 \scriptstyle{\pm 0.10}$    &       &   $6$    &  $9.59 \scriptstyle{\pm 0.67}$ \\
    DiT + Min-Max + DSR  & $\mathbf{23}$  & $\mathbf{6.62 \scriptstyle{\pm 0.20}}$  &       & $\mathbf{45}$ &  $\mathbf{14.44 \scriptstyle{\pm 0.40}}$\\
    \bottomrule
    \end{tabular}%

}
\label{tab:main}
\end{table}


\begin{table}[!t]
\centering
\caption{Ablation study on min-max losses, the result is tested on ImageNet-tiny with only 1 training epoch. Bold indicates the best.}

\renewcommand{\arraystretch}{1.5}
\resizebox{.35\textwidth}{!}{
    \begin{tabular}{ccccccccc}
    \toprule
    \textbf{$\mathbf{\mathcal{L}_{diff}}$}& & \cmark  &   & \cmark  &   & \cmark   &  & \cmark \\
    \textbf{$\mathbf{\mathcal{L}_{r}}$}& &   &    & \cmark   &  &  &     & \cmark \\
    \textbf{$\mathbf{\mathcal{L}_{d}}$}& &   &    &   &    & \cmark   &  & \cmark \\
    \midrule
    \midrule
    Accuracy& & 1.30 &  & \textbf{1.43} & & 1.10 &  & 1.41 \\
    \bottomrule
    \end{tabular}%

}
\label{tab:ablation}
\end{table}

\begin{table}[!t]
\centering
\caption{Ablation on the number of diffusion steps of RSD, showing the trade-off of the image quantitative and quality in the surrogate dataset. Higher accuracy gain indicates better image quality. Bold indicates the best.}
\renewcommand{\arraystretch}{1.5}
\resizebox{.85\textwidth}{!}{

    \begin{tabular}{ccccccccc}
    \toprule
    \multirow{2}[4]{*}{\textbf{Diff. Steps}} & \multirow{2}[4]{*}{Batchtime (s)} & \multirow{2}[4]{*}{\# of Samples} &       & \multicolumn{2}{c}{ImageNet-tiny} &       & \multicolumn{2}{c}{CIFAR-100} \\
\cmidrule{5-6}\cmidrule{8-9}          &       &       &       & Accuracy & Acc. Gain &       & Accuracy & Acc. Gain \\
    \midrule
    5 & \textbf{2.5}   & \textbf{4544}  &       & 5.39  & 1.19  &       & 11.98 & 2.64 \\
    10 & 5     & 2560  &       & \textbf{6.53}  & 2.55  &       & \textbf{14.72} & 5.75 \\
    15 & 7.5   & 1792  &       & 5.88  & 3.28  &       & 13.86 & 7.73 \\
    20 & 10    & 1376  &       & 5.24  & 3.81  &       & 12.44 & 9.04 \\
    25 & 12.5  & 1120  &       & 4.88  & 4.36  &       & 13.32 & 11.89 \\
    30 & 15    & 800   &       & 4.19  & \textbf{5.24}  &       & 11.01 & \textbf{13.76} \\
    \bottomrule
    \end{tabular}%

   }
\label{tab:diffstep}
\end{table}

\subsection{Overall Results}
\label{quantitative}
In Table~\ref{tab:main}, our method demonstrates superior performance over the baseline, with improvements of $89\%$ on ImageNet-tiny and $156\%$ on CIFAR-100. The incorporation of Diffusion Steps Reduction (DSR) also significantly enhances image generation speed, resulting in a higher IPC. Figure~\ref{exp1} showcases qualitatively generated examples from the ImageNet-tiny and CIFAR-100 datasets. Even with only 10 diffusion steps, the generated images effectively represent their respective classes by capturing key features, such as dogs and vehicles. Although image quality decreases with fewer diffusion iterations, the critical features remain sufficient for classification tasks. Notably, increasing to 30 diffusion steps would improve image quality but slow down the image generation speed.

\subsection{Ablation Studies}
\label{qualitative}
In Table~\ref{tab:ablation}, we present an ablation study on the two min-max losses. We find that the representativeness loss $\mathcal{L}_r$ improves performance, while the diversity loss $\mathcal{L}_d$ negatively impacts it. We argue that under the time constraints of generative dataset distillation, the quantity of the generated images is usually insufficient, limiting the diversity of the surrogate dataset, and making representativeness more important than diversity. Therefore, we recommend prioritizing representativeness loss when generative models struggle to produce sufficient images within time constraints.

In Table~\ref{tab:diffstep}, we analyze how the number of diffusion steps in DSR affects the quantity of generated images (number of samples), accuracy, and accuracy gain across two datasets. Accuracy gain, calculated as $Acc. Gain = \frac{Accuracy}{\# of samples}$, is used to assess the quality of the generated images. Our results show that increasing diffusion steps improves image quality but reduces the number of generated samples, highlighting the trade-off between image quality and quantity. We determine that setting the diffusion step to 10 offers the optimal balance for our approach.

\section{Conclusion}\label{sec:conclusion}
This paper proposes the diffusion-based method for dataset distillation using a generative model. We leverage min-max training losses to ensure the representativeness and diversity of the surrogate dataset. Furthermore, we propose a strategy to increase the number of generated images by reducing the diffusion steps. Our model achieved $2^{nd}$ place in the challenge, demonstrating its superior performance. In the future, we will develop more effective distillation techniques with generative models across different datasets.

%
%
\bibliographystyle{splncs04}
\bibliography{main}
\end{document}